\documentclass[11pt,a4paper]{article}
\usepackage{authblk}
\usepackage[hyperref]{emnlp2020}
\usepackage{times}
\usepackage{latexsym}

\usepackage{adjustbox}
\usepackage{url}
\usepackage{amsmath}
\usepackage{multirow}
\usepackage{tabularx}
\usepackage{tablefootnote}

\usepackage{microtype}

\aclfinalcopy 

\newcommand{\hidden}[1]{}

\title{Analyzing Neural Discourse Coherence Models}

\author[1,2]{\textbf{Youmna Farag}}
\author[1]{\textbf{Josef Valvoda}}
\author[3]{\textbf{Helen Yannakoudakis}}
\author[1,2]{\textbf{Ted Briscoe}}
\affil[1]{Department of Computer Science and Technology, University of Cambridge, United Kingdom}
\affil[2]{The ALTA Institute, Cambridge, United Kingdom}
\affil[ ]{\tt \{youmna.farag,jv406,ted.briscoe\}@cl.cam.ac.uk}
\affil[3]{Department of Informatics, King’s College London, United Kingdom}
\affil[ ]{\tt helen.yannakoudakis@kcl.ac.uk}

\date{}

\begin{document}
\maketitle

\begin{abstract}

In this work, we systematically investigate how well current models of coherence can capture aspects of text implicated in discourse organisation. We devise two datasets of various linguistic alterations that undermine coherence and test model sensitivity to changes in syntax and semantics. We furthermore probe discourse embedding space and examine the knowledge that is encoded in representations of coherence. We hope this study shall provide further insight into how to frame the task and improve models of coherence assessment further. Finally, we make our datasets publicly available as a resource for researchers to use to test discourse coherence models.  

\end{abstract}

\section{Introduction}
Coherence refers to the properties of a text that indicate how meaningful (sub-)sentential constituents are connected to convey document-level meaning. Different theories have been proposed to describe the properties that contribute to discourse coherence and some have been integrated with computational models for empirical evaluation. A popular approach is the entity-based model which hypothesizes that coherence can be assessed in terms of the distribution of and transitions between entities in a text -- by constructing an entity-grid (Egrid) representation~\cite{Barzilay:2005,Barzilay2008}, building on Centering Theory~\cite{Grosz1995}. Subsequent work has adapted and further extended Egrid representations~\cite{filippova-strube-2007-extending,Burstein2010,Elsner2011,Guinaudeau2013}. 
Other research has focused on syntactic patterns that co-occur in text~\cite{louis-nenkova-2012-coherence} or semantic relatedness between sentences~\cite{Lapata:2005:AET:1642293.1642467,soricut-marcu-2006-discourse,somasundaran-etal-2014-lexical} as key aspects of coherence modeling.
There have also been attempts to model coherence by identifying rhetorical relations that connect textual units~\cite{mann1988rhetorical,lin-etal-2011-automatically,Feng2014} or capturing topic shifts via Hidden Markov Models~\cite[HMM,][]{barzilay-lee-2004-catching}. Other work has combined approaches to study whether they are complementary~\cite{elsner-etal-2007-unified,Feng2014}. 
More recently, neural networks have been used to model coherence. Some models utilize structured representations of text~\cite[e.g. Egrid representations,][]{Dat2017,Joty2018} and others operate on unstructured text, taking advantage of neural models' ability to learn useful representations for the task~\cite{Li2017,Logeswaran2018,farag-yannakoudakis-2019-multi,xu-etal-2019-cross,moon-etal-2019-unified}.

Coherence has typically been assessed by a model's ability to rank a well-organized document higher than its noisy counterparts created by corrupting sentence order in the original document (\textit{binary discrimination task}), and neural models have achieved remarkable accuracy on this task. Recent efforts have targeted additional tasks such as recovering the correct sentence order~\cite{Logeswaran2018,Cui2018}, evaluating on realistic data~\cite{Lai2018,farag-yannakoudakis-2019-multi} and focusing on open-domain models of coherence~\cite{Li2017,xu-etal-2019-cross}.
However, less attention has been directed to investigating and analyzing the properties of coherence that current models can capture, nor what knowledge is encoded in their representations and how it might relate to aspects of coherence. 

In this work, we systematically examine what properties of discourse coherence current coherence models can capture. We devise two datasets that exhibit various kinds of incoherence and analyze model ability to capture syntactic and semantic aspects of text implicated in discourse organisation. 
We furthermore investigate a set of probing tasks to better understand the information that is encoded in their representations and how it might relate to aspects of coherence. 
We hope this study shall provide further insight into how to frame the task and improve models of coherence assessment further. 
Finally, we release our evaluation datasets as a resource for the community to use to test discourse coherence models.\footnote{\url{https://github.com/Youmna-H/coherence-analysis}} 

\section{Neural Coherence Models}
\label{models}
We experiment with a number of existing and state-of-the-art neural approaches to coherence assessment, that have publicly available implementations, and present details of the models below. Across all the BERT-based models, we use bert-large-uncased and layer $16$ following \citet{liu2019linguistic} and~\citet{hewitt2019structural}.

\noindent{\bf Multi-task learning}~\cite[\textbf{MTL},][]{farag-yannakoudakis-2019-multi}: The model applies a Bi-LSTM on input GloVe word embeddings~\cite{pennington2014glove} followed by attention to build sentence representations; then builds a second Bi-LSTM with attention to compose a document vector. A linear operation followed by a sigmoid function is applied to the document representation to predict an overall coherence score as the main objective. Inspired by the Egrid approaches, the model is also optimized to predict the grammatical roles of the input words at the bottom layer of the network as an auxiliary task. 

\noindent{\bf MTL with BERT embeddings (MTL$_{bert}$)}: We replicate the previous MTL model but now use BERT embeddings \cite{devlin2018bert} to initialize the input words. 

\noindent{\bf Single-task learning}~\cite[\textbf{STL},][]{farag-yannakoudakis-2019-multi}: This model has the same architecture as MTL but only performs the coherence prediction task, excluding the grammatical role auxiliary objective.  

\noindent{\bf STL with BERT (STL$_{bert}$)}: This is the same as STL but uses BERT embeddings.

\noindent{\bf Local Coherence Discriminator with Language modeling}~\cite[\textbf{LCD$_{rnnlm}$},][]{xu-etal-2019-cross}: The model generates sentence representations via an RNN language model, where word embeddings are initialized using GloVe.
It then generates a representation for two consecutive sentences via concatenating the output of a set of linear transformations applied to the two sentences: concatenation, element-wise difference, element-wise product and absolute value of element-wise difference. This representation is fed to an MLP layer to predict a \textit{local} coherence score.\footnote{Gold local scores $\in \{0,1\}$ represent whether a sequence of two sentences is coherent (i.e. extracted from a coherent document) or not (i.e. created via negative sampling).} 
The overall coherence of a document is the average of its local scores.

\noindent{\bf LCD with BERT (LCD$_{bert}$)}: We create a variant of the LCD$_{rnnlm}$ model where instead of using an RNN language model encoder, we encode each sentence as the average BERT vectors of the words it contains. Everything else remains the same.

\noindent{\bf Local Coherence}~\cite[LC,][]{Li2017}: The model generates sentence vectors via an LSTM over GloVe-initialized word embeddings; then a window approach is applied over adjacent sentences to get embeddings of groups of sentences and predict local coherence scores. The final document score is calculated by averaging its local scores.

\noindent{\bf Egrid CNN}~\cite[Egrid$_{cnn}$,][]{Dat2017}: The model applies a CNN over Egrid representations across groups of consecutive sentences; the CNN slides multiple filters of weights to extract feature maps that represent high-level entity-transition features, followed by a max pooling function to focus on the important features. Furthermore, additional entity-related features are integrated such as salience, proper mentions and named entity type.

\begin{table*}[]
\centering
\scalebox{0.7}{
\begin{tabular}{|cccccccc|} \hline
 MTL   & MTL$_{bert}$ & STL   & STL$_{bert}$ &  LCD$_{rnnlm}$ & LCD$_{bert}$ & LC    & Egrid$_{cnn}$                 \\ \hline
93.2 & 96.1 & 87.7 & 95.4 & 94.5 & \textbf{97.1} & 74.1 &  87.6 \\ \hline
\end{tabular}}
\caption{PRA results of coherence models based on the binary discrimination task on the WSJ.} 
\label{table1}
\end{table*}

\section{Binary Discrimination Task}
\label{bintask}
Binary discrimination is a typical approach to assessing neural coherence models where a well-organized document should be ranked higher than its permuted counterparts created by corrupting sentence order. Following previous work, we train and test\footnote{All models are run $5$ times and the test predictions are averaged across the runs.} the coherence models on the WSJ\footnote{We use the same train and test splits as~\citet{Dat2017} and the same test set permuted counterparts as~\citet{farag-yannakoudakis-2019-multi}.} and  evaluate them using Pairwise Ranking Accuracy (PRA), which is calculated based on the fraction of correct pairwise rankings between a coherent document and its incoherent counterparts. 

In Table ~\ref{table1}, we present the performance of all coherence models. The high accuracy of the models demonstrates their efficacy for the task of selecting a maximally coherent sentence order from a set of candidate permutations. We note that the LCD and MTL BERT variants achieve a new state-of-the-art on the WSJ. The remarkable accuracy on this task may render this problem fully solved.

Herein, we seek to investigate how well these models of coherence can capture aspects of text implicated in discourse organisation. We devise a set of datasets and systematically test model susceptibility to syntactic or semantic changes. 

\begin{table*}[]
\centering
\scalebox{0.7}{
\begin{tabular}{|l|l|}
\hline
\multicolumn{1}{|c|}{Type of reference} & \multicolumn{1}{c|}{Example}                                                                                                                                            \\ \hline
Pronominal Reference                    & Rich was a musician. \underline{He} made a few hit songs.                                                                                                                           \\ \hline
Proper Name                             & Dan's parents were overweight. \underline{Dan} was overweight as well.                                                                                                              \\ \hline
Nominal Substitution                    & My dog hates his treats. I decided to go buy some new  \underline{ones}.                                                                                                             \\ \hline
Demonstrative Reference                 & \begin{tabular}[c]{@{}l@{}}My daughter wants to take her toddler to the Enchanted Village.\\  \underline{This} is a puppet show featuring early 20th century figurines.\end{tabular} \\ \hline
\end{tabular}}
\caption{Examples of first two sentences extracted from the ROCStories Cloze dataset with different referential types (referring word underlined).}
\label{table2}
\end{table*}

\section{Cloze Coherence (CC) Dataset }
\label{ch4:directevaluation}

We compile a large-scale dataset, to which we refer as Cloze Coherence (CC), of coherent and incoherent examples, where the former are intact well-written texts while the latter are the result of applying syntactic or semantic perturbations to the coherent ones. 

\subsection{Coherent examples}
For the sake of specifically testing for coherence, we avoid complex linguistic structures. Specifically, we focus on coherent examples that consist of two short sentences that are coreferential and exhibit a rhetorical relation (such properties can be manipulated to create incoherent counterparts). Furthermore, we focus on examples that are self-contained, meaning that they do not reference or rely on an outer context to be interpreted. We find that narrative texts are good candidates to satisfy these criteria and therefore create our coherent examples from the ROCStories Cloze dataset\footnote{\url{https://www.cs.rochester.edu/nlp/rocstories/}}~\cite{mostafazadeh-etal-2016-corpus}.

ROCStories Cloze contains short stories of $5$ sentences manifesting a sequence of causal or temporal events that have a shared protagonist. A story usually starts by introducing a protagonist in the first sentence, then subsequent sentences describe events that happen to them in a logical / rhetorically plausible manner. The dataset was designed for commonsense reasoning by testing the ability of machine learning models to select a plausible ending for the story out of two alternative endings. Here, our main aim is to challenge the models and investigate whether they truly understand 
inter-sentential relations and coherence-related features. We specifically utilize the first two sentences in the stories to compose the coherent examples in our dataset.\footnote{We use NLTK for word tokenization; sentence boundaries are already marked in the stories.}

Selecting the first two sentences helps make the examples self-contained since there is no preceding context to refer to, and no cataphoric relations to consequent sentences. Regarding rhetorical relations in these sentences,~\citet{mostafazadeh-etal-2016-corpus} conducted a temporal analysis to investigate the logical order of the events presented in a story, demonstrating, among others, that the first and second sentences in the stories are presented in a commonsensical temporal manner with logical links between them.  
In order to examine coreferential relations between the two sentences in each extracted pair, we gather a set of statistics. We adopt a heuristic approach\footnote{We initially used the spaCy and Stanford coreference resolution systems~\cite{clark2016deep}, but found their performance unreliable for the purposes of this experiment after manual inspection.} by simply counting the number of second sentences that contain at least one third person pronoun (either personal or possessive) and find that they constitute $80\%$ of the examples.\footnote{If we exclude `it' the percentage becomes $76\%$.} Third person pronouns anaphorically refers to preceding items in text, which could occur in the same sentence or the previous one (i.e., the first sentence). We, therefore, randomly select, and manually inspect, $500$ examples that contain third person pronouns in their second sentence and find that in $95\%$ of them the referenced entity appears in the first sentence. Furthermore, third person pronouns are not the only coreferential relations in the examples. For instance, we find that $90\%$ of the second sentences contain a personal or possessive pronoun (whether it is first, second or third person), which could also signal coreference, e.g., `\textit{I was walking to school. Since I wasn't looking at my feet I stepped on a rock.}'  There are also other coreferential devices such as: demonstrative references (e.g., ‘this’ and ‘there’), ‘the’ + noun, proper names or nominal substitutions (e.g., ‘one’ or ‘ones’) to name a few~\cite{halliday}, so the true proportion of coreferential pairs will be higher. Table~\ref{table2} presents examples of different referential relations in our dataset. 

We use the same train/dev/test splits provided with ROCStories Cloze but only keep the first two sentences in each story. We exclude cases with erroneous sentence boundaries,\footnote{The training stories are in CSV format (separating sentences by comma delimiters) and we parse them using the Python CSV parser. We exclude the stories where the parser fails to detect $5$ sentences.} yielding $97,903$ examples for training, $1,871$ for development, and $1,871$ for testing, and a training vocabulary size of $29,596$ tokens. Each instance in our dataset contains two sentences that represent a coherent pair.

\begin{table*}[ht] 
\centering
\scalebox{0.9}{
\small
\begin{tabularx}{\textwidth}{ 
  | >{\raggedright}X 
  | >{\raggedright}X 
  | >{\raggedright\arraybackslash}X | }
\hline
\multicolumn{1}{|c|}{Coherent example}   & \multicolumn{1}{c|}{Incoherent example from cloze\_swap} & \multicolumn{1}{c|}{Incoherent example from cloze\_rand}
\\ \hline
Tyrese joined a new gym.The membership allows him to work out for a year.                  & The membership allows him to work out for a year. Tyrese joined a new gym. & Tyrese joined a new gym. As children they hated being dressed alike.                     \\ \hline
Jasmine doesn't know how to play the guitar. She asked her dad to take her to guitar class.  & She asked her dad to take her to guitar class. Jasmine doesn't know how to play the guitar.  & Jasmine doesn't know how to play the guitar. May thought her milk was no good.           \\ \hline
I wanted to play an old game one day. When I looked in the game's case the CD was missing. & When I looked in the game's case the CD was missing. I wanted to play an old game one day.  & I wanted to play an old game one day. Jason pressed the buzzer since he knew the answer. \\ \hline
\end{tabularx}}
\caption{Examples of coherent and incoherent pairs from the cloze\_swap and cloze\_rand datasets.}
\label{table3}
\end{table*}

\subsection{Incoherent examples}
To assess model susceptibility to syntactic or semantic alterations, we construct incoherent examples by applying two different transformations to each coherent pair resulting in two different sets of data.

\noindent{\bf cloze\_swap} We create incoherent examples by swapping the two sentences in a coherent pair. This mostly breaks the coreference relation between them and/or the rhetorical relation (e.g. temporal or causal) by reversing the event sequence. The dataset, referred to as \texttt{cloze\_swap}, is balanced, i.e., the number of incoherent examples is the same as the number of the coherent ones above. The way cloze\_swap is created corrupts the syntactic patterns that co-occur in coherent texts (e.g. $\text{S} \rightarrow \text{NP-SBJ VP } | \text{ NP-SBJ} \rightarrow \text{PRP}$) as demonstrated by~\citet{louis-nenkova-2012-coherence}.  

\noindent{\bf cloze\_rand} Here we create incoherent examples by keeping the first sentence of a coherent pair intact and replacing the second with a randomly selected second sentence from (the same split of) our set of coherent examples. 
This dataset, referred to as \texttt{cloze\_rand}, is also balanced (for each coherent pair, we compose one incoherent counterpart), and constitutes examples with changed semantics but with the main syntactic pattern intact. 
As the randomly-created pair may still be coherent, we address this by: 1) constraining random selection of the second sentence to not begin with the same word as the second sentence in the original pair, or with the pronoun `he' if the original starts with `she', and vice-versa\footnote{We do not find instances of `they' as a third-person singular pronoun.} (we note $70\%$ of the second sentences in ROCStories Cloze start with a pronoun); 2) using human evaluation to further assess the validity of this data and get an estimate of upper-bound performance on the task. Specifically, we randomly select $100$ coherent sentence pairs from our test split along with their own incoherent counterparts and ask two annotators (who are not authors of this paper), with high English proficiency levels, to \textit{rank} each set of coherent--incoherent examples based on which one they considered to be more coherent and plausible. The average PRA of the annotators is $94.5\%$.

Table~\ref{table3} shows examples from cloze\_swap and cloze\_rand. As our datasets are balanced (one incoherent counterpart per coherent pair), we have a total number of $195,806$, $3,742$ and $3,742$ instances in the train, dev, and test splits respectively for each cloze dataset (cloze\_swap and cloze\_rand have the same coherent examples, and the same number of coherent and incoherent examples).

We note that the gold labels in this data are not to be interpreted as (overall) binary indicators of coherence. We rather use these to test model performance using PRA, i.e. we only compare a coherent pair with \textit{its own} incoherent counterpart.

\begin{table*}[]
\centering
\scalebox{0.55}{
\begin{tabular}{|c|l|} \hline
Original              & \begin{tabular}[c]{@{}l@{}}A government paper on Monday found UK and EU firms would be faced with a "a significant new and ongoing administrative burden" in the event of a no-deal Brexit.\\ It found large firms importing and exporting at scale would need to fill in forms taking one hour 45 minutes on average and cost £28 per form for each load imported.\end{tabular}                                                                                                                                                                                                            \\ \hline
Swap                  & \begin{tabular}[c]{@{}l@{}}It found large firms importing and exporting at scale would need to fill in forms taking one hour 45 minutes on average and cost £28 per form for each load imported.\\ A government paper on Monday found UK and EU firms would be faced with a "a significant new and ongoing administrative burden" in the event of a no-deal Brexit.\end{tabular}                                                                                                                                                                                                            \\ \hline
Random                & \begin{tabular}[c]{@{}l@{}}1- A government paper on Monday found UK and EU firms would be faced with a "a significant new and ongoing administrative burden" in the event of a no-deal Brexit.\\ She spent over a decade at Swiss investment bank UBS before joining the UK Treasury's council of economic advisers in 1999.\\ 2- Lady Vadera was born in Uganda and moved to the UK as a teenager.\\ It found large firms importing and exporting at scale would need to fill in forms taking one hour 45 minutes on average and cost £28 per form for each load imported.\end{tabular} \\ \hline
\begin{tabular}[c]{@{}c@{}}Lexical\\ Substitution\end{tabular} & \begin{tabular}[c]{@{}l@{}}The paper found large firms importing and exporting at scale would need to fill in forms taking one hour 45 minutes on average and cost £28 per form for\\each load imported.\\ A government paper on Monday found UK and EU firms would be faced with a "a significant new and ongoing administrative burden" in the event of a no-deal Brexit.\end{tabular}                                                                                                                                                                                                     \\ \hline
\begin{tabular}[c]{@{}c@{}}Prefix\\ Insertion\end{tabular}      & \begin{tabular}[c]{@{}l@{}}More Specifically, it found large firms importing and exporting at scale would need to fill in forms taking one hour 45 minutes on average and cost £28 per form\\ for each load imported.\\ A government paper on Monday found UK and EU firms would be faced with a "a significant new and ongoing administrative burden" in the event of a no-deal Brexit.\end{tabular}                                                                                                                                                                                        \\ \hline
\begin{tabular}[c]{@{}c@{}}Lexical\\ Perturbations\end{tabular} & \begin{tabular}[c]{@{}l@{}}A government paper on Monday found UK and EU firms would be faced with a "a significant new and ongoing administrative burden" in the event of a no-deal Brexit.\\ It found large firms importing and exporting at scale would need to fill in cups taking one hour 45 minutes on average and cost £28 per cup for each load imported.\end{tabular}                                                                                     \\ \hline
\begin{tabular}[c]{@{}c@{}}Corrupt\\ Pronoun\end{tabular}   & 
\begin{tabular}[c]{@{}l@{}}A government paper on Monday found UK and EU firms would be faced with a "a significant new and ongoing administrative burden" in the event of a no-deal Brexit.\\He found large firms importing and exporting at scale would need to fill in forms taking one hour 45 minutes on average and cost £28 per form for each load imported.\end{tabular}                                                                                                                                                                                                            \\ \hline
\end{tabular}}
\caption{Examples from our manually constructed CLA dataset. For `Random' we create two incoherent instances: one where the first sentence is unchanged and the second is randomly selected (1-); and another where the first sentence is randomly selected and the second is kept intact (2-).} 
\label{table5}
\end{table*}

\section{Controlled Linguistic Alterations (CLA) Dataset }

In order to further understand the properties of coherence that current coherence models capture, we manually construct a dataset of controlled sets of linguistic changes. We first identify a set of coherent, well-written texts of two consecutive sentences from business and financial articles in the BBC, the Independent and Financial Times (this allows us to stay in the same domain as the one used for training the models -- the WSJ). We focus on sentence pairs where the subject of the first sentence is pronominalized in the second, and the second sentence begins with this pronoun. We select the examples so that they are self-contained and do not reference an outer context. 
We then manually create incoherent counterparts by modifying the coherent examples in a constrained way in order to systematically examine model performance. Specifically, we apply the following sets of perturbations to our set of coherent sentence pairs, examples of which are presented in Table~\ref{table5}.

\noindent{\bf{Swap.}} We simply swap the two sentences. 
\vspace{0.1cm} \\ 
\noindent{\bf{Random.}} We keep the first sentence intact and select a second sentence randomly from our set of coherent examples. We constrain the selection so that the subject pronoun is different from the subject pronoun in the original sentence.\footnote{We also take into account that some subjects could be referred to by `he', `she' or `they' and thus factor that into the selection.} We also create another random pair with the same constraint but now changing the first sentence. Thus each original coherent example has two incoherent counterparts. 
\vspace{0.1cm} \\ 
\noindent{\bf{Lexical Substitution.}} We swap the two sentences in a coherent pair but replace the subject pronoun in the second sentence with \textit{the + a general noun} that substitutes the subject in the first sentence (e.g. the company, the woman, etc.). 
\vspace{0.1cm} \\ 
\noindent{\bf{Prefix Insertion.}} We analyze the WSJ training data and find that the average number of times the first sentence in a document starts with a pronoun is $0.02$ (and never with `he' or `she') which is significantly less than the average number of times a sentence starts with a pronoun (regardless of its position) which is $0.07$. This difference is not maintained in the randomly ordered documents in the WSJ training set and so this might give a signal to the models to detect that a swapped pair that starts with a pronoun is less coherent. To see if such positional information plays a role in model prediction, we insert a phrase, before the subject pronoun after swapping the sentences, that doesn't change the propositional content (e.g. `More specifically', `However', etc.). We can then observe whether this insertion will change the prediction of the model.
\vspace{0.1cm} \\ 
\noindent{\bf{Lexical Perturbation.}} We investigate the robustness of the models to minor lexical changes that result in incoherent meaning, by replacing one word in either of the two sentences (if the word is repeated, we change that too). We choose a replacement word from the training vocabulary of the WSJ with the same part-of-speech tag. For example, in Table~\ref{table5} `form' is replaced with `cup' and `forms' with `cups'.
\vspace{0.1cm} \\ 
\noindent{\bf{Corrupt Pronoun.}} We replace the subject pronoun in the second sentence with another pronoun that cannot reference anything in the first sentence. With this method, we test whether the models are capable of resolving coreferences or just rely on syntactic patterns.

Our dataset contains a total of $240$ examples of coherent and incoherent pairs of sentences ($30$ coherent examples and $210$ incoherent counterparts). 
Our constrained set of modifications ensures that all coherent examples are more coherent than any of the incoherent counterparts in the data.    

\begin{table*}[]
\centering
\scalebox{0.7}{
\begin{tabular}{|c|c|c|cccccccc|} \hline
  & \multirow{2}{*}{Dataset} & \multirow{2}{*}{\# comparisons} & \multicolumn{8}{c|}{Models}                                                                                   \\
                   &                        &                                 & MTL   & MTL$_{bert}$ & STL   & STL$_{bert}$ & LC    & LCD$_{rnnlm}$ & LCD$_{bert}$ & Egrid$_{cnn}$                 \\ \hline
\multirow{4}{*}{CC}  & cloze\_swap & 1,871 & 69.3 & 73.5 & 74.2  & 75.3  & 70.7  & 74.5  & 75.4  &  \textbf{84.6} \\ 
                     & fine-tuned & 1,871 & 88.8 & 88.5 & 83.5  & 84.7  & 76.3  & 88.4  & \textbf{96.7}  & 88.1  \\ \cline{2-11}
                     & cloze\_rand & 1,871 & 51.3 & 53.3 & 48.5 & 52.5 & 50.5 & 54.5 & \textbf{71.0} & 53.4 \\ 
                     & fine-tuned & 1,871 & 65.7 & 54.2 & 53.7  & 56.1 & 51.3 & 65.2 & \textbf{94.8} & 68.8 \\ \hline \hline

\multirow{8}{*}{CLA} &    \multicolumn{1}{c|}{ Swap}                  & \multicolumn{1}{c|}{30}         & 90.0   & \textbf{93.3}    & 83.3 & 90.0      & 80.0   & \textbf{93.3} & 86.6    &       83.3                      \\
&    \multicolumn{1}{c|}{Random}                & \multicolumn{1}{c|}{60}         & 56.6 & 45.0    & 50.0   & 51.6    & 51.6 & 61.6 & \textbf{78.3}    &    71.6                             \\
&    \multicolumn{1}{c|}{Lexical Substitution}  & \multicolumn{1}{c|}{30}         & 83.3 & \textbf{93.3}    & 80.0   & 90.0      & 86.6 & 83.3 & 86.6      &       76.6                           \\
&    \multicolumn{1}{c|}{Prefix Insertion}      & \multicolumn{1}{c|}{30}         & 83.3 & \textbf{96.6}    & 76.6 &90.0    & 76.6 & 86.6 & 93.3    &      80.0                         \\
&    \multicolumn{1}{c|}{Lexical Perturbations} & \multicolumn{1}{c|}{30}         & 56.6 & 46.6      & 46.6 & 63.3    & 50.0   & 53.3 & \textbf{80.0}    &     53.3                              \\
&    \multicolumn{1}{c|}{Corrupt Pronoun}       & \multicolumn{1}{c|}{30}         & 70.0   & 53.3    & 63.3 & 63.3    & 53.3 & 60.0   & \textbf{76.6}    &      56.6                             \\ 
\cline{2-11} &    \multicolumn{1}{c|}{All data}                   & \multicolumn{1}{c|}{210}        & 70.9 & 67.6    & 64.2 & 71.4    & 64.2 & 71.4 & \textbf{82.8}    &          70.4                  \\ 
\cline{2-11}&    \multicolumn{1}{c|}{All data (TPRA)}                  & \multicolumn{1}{c|}{6,300}     & 69.9 & 71.3    & 61.8 & 71.6    & 66.0  & 69.1 & \textbf{72.2}    & 65.8  \\ \hline

\end{tabular}}
\caption{PRA performance on the CLA (bottom) and CC datasets (top; `fine-tuned' shows results for models tuned on the respective cloze training sets).}
\label{table6}
\end{table*}

\section{Experiments}
Table \ref{table6} (top) presents the PRA performance of the models trained on the WSJ (Section \ref{bintask}) when they are evaluated on the test sets of the CC datasets (rows `cloze\_swap' and `cloze\_rand'). We find that, overall, models are good at detecting syntactic alterations (cloze\_swap; PRA ranging from $69.3$ to $84.6$) even though the test data is from a domain different than the training one. 
 However, most models perform poorly on semantic alterations (cloze\_rand; PRA ranging from $48.5$ to $54.5$), the only exception being LCD$_{bert}$ that achieves a PRA of $71$. Specifically, models that use RNN-based sentence encoders (the first six models), even when initialised with BERT, or apply a CNN to capture entity transitions fall short in capturing semantic changes despite the fact that cloze\_rand is from the same domain as cloze\_swap. In contrast, LCD$_{bert}$ is more capable of detecting semantic changes where the model builds sentence representations by averaging BERT vectors then applies a set of linear transformations to increase its expressive power, surpassing its RNN-based counterpart (LCD$\_{rnnlm}$) with $16.5$\% on cloze\_rand. Additionally, across models, we observe that the use of contextualized (BERT) embeddings consistently improves performance on both cloze tasks, although performance on semantic alterations remains close to random.  
 
We investigate domain shift effects and fine-tune the WSJ-trained models on each of the cloze\_swap and cloze\_rand training sets (Section \ref{ch4:directevaluation}) and re-evaluate performance on the respective test sets. Specifically, we use  an MLP layer over the models' pre-prediction representation, followed by sigmoid non-linearity. The models are optimized using the mean squared error between the gold labels (0 or 1) and the predicted scores.\footnote{We use Adam~\citep{KingmaB14}, batch size $64$, L$2$ regularization, and a learnable penalty rate (search space $\{0.00001, 0.0001, 0.001, 0.01\}$). We use early stopping and stop training if PRA does not improve on the dev set over $5$ epochs (max epochs $200$). MLP hidden unit size is $100$.} In this setup, only the MLP layer is fine-tuned and not the whole coherence model which allows us to create a fast efficient evaluation framework that can be applied as a further examination step after coherence models are developed and tuned on their respective datasets, instead of training the models from scratch. The results of the fine-tuned models are presented in Table~\ref{table6} (CC; rows `fine-tuned'). Although we can see that there is some domain effect, we nevertheless find that the results confirm our earlier observation: performance on semantic alterations is, overall, poor, in contrast to syntactic ones (cloze\_swap).    

In Table \ref{table6} (bottom), we can observe model performance (PRA) on our constrained set of manually devised examples (CLA). Again, we observe a similar result: across RNN-based models, performance is particularly low on random examples, which suggests that they struggle to detect topical or rhetorical shifts and unresolved references if the main syntactic pattern is maintained. 
The exception is LCD$\_{bert}$ which is again the best performing model (PRA $78.3$). 

We furthermore observe that now Egrid$_{cnn}$ is the second best model on CLA Random (PRA $71.6$). A sparser entity grid where entities in the two sentences are different allows the model to detect such cases (e.g.~in the example in Table~\ref{table5}, `firms' is mentioned in the two sentences, while in the two random examples, it is only mentioned in one). However, its substantial difference in PRA on CLA Random compared to cloze\_rand ($53.4$) suggests that the lower performance observed in the latter is due to domain shift effects, something which we do not observe (to the same extent) with LCD$_{bert}$. 
Regarding the CLA Swap results, we can again confirm models' capability of detecting corrupted syntactic constructions. We furthermore observe that they are able to maintain good performance in the cases where a prefix is inserted (`Prefix Insertion') or the subject pronoun is substituted with a lexical item (`Lexical Substitution'). This suggests that they can capture the relevant syntactic patterns and do not rely solely on positional features. 

\begin{table*}[]
\centering
\scalebox{0.75}{

\begin{tabular}{|c|cccccccl|c|} \hline

\multirow{2}{*}{Task}              & \multicolumn{8}{c|}{Models} &     \multirow{2}{*}{Human}                                                                                                                                                             \\ 
            & MTL   & MTL$_{bert}$  & STL   & STL$_{bert}$  & LC  &  LCD$_{rnnlm}$ & LCD$_{bert}$ & Best from~\citet{conneau-etal-2018-cram} &   \\ \hline
SubjNum     & 64.9  & 75.4          & 62.2  & 71.5          & 52.7             & 71.2         & \textbf{88.0} &                  95.1 (Seq2Tree)& 88.0                                                      \\ \hline
ObjNum       & 64.5   & 72.1          & 61.1  & 70.7          & 54.5 &   65.0 & \textbf{86.5} &               95.1 (Seq2Tree) &86.5                                                      \\ \hline
CoordInv    & 58.5   & 63.4          & 53.0    & {63.7} & 53.0 &   56.6 & \textbf{78.4}           &                    76.2 (NMT En-De) &85.0                                                    \\ \hline
CorruptAgr  & 53.2 & 69.7 & 57.7 & 68.6           & 52.2   &  64.2 &   \textbf{94.3}             &                    -       & -                                    \\ \hline
\end{tabular}
}
\caption{Classification accuracy on probing tasks. `Human' shows the human upper bound on the task. } 
\label{table4}
\end{table*}

Performance is overall low on lexical perturbations and corrupt pronouns which suggests that the models are not sensitive to minor lexical changes even if they result in implausible meaning and they also struggle to resolve pronominal references. However, the exception is LCD$_{bert}$ (with PRA $80$ on lexical perturbations and $76.6$ on corrupt pronoun) suggesting a better ability at capturing semantics and resolving references.

Across all six CLA datasets (`All data'; Table \ref{table6}), we find that, overall, LCD$_{bert}$ is the top performing model (average PRA). The `All data' row reports the result of comparing a coherent example against its incoherent counterparts across the different alterations (i.e., in Table~\ref{table5}, the original example is compared against all the examples in the table and this is applied to all the original examples in the dataset). If we furthermore compare all the coherent examples against the incoherent ones in the whole dataset (rather than against their own incoherent counterparts), we find that a similar performance pattern is maintained (row `All data (TPRA)', i.e., all data Total Pairwise Ranking Accuracy). 

\section{Probing Coherence Embedding Space}
\label{probe} 

Inspired by previous work ~\cite{conneau-etal-2018-cram}, and to better understand the information that is encoded in the representations of coherence models, we investigate probing tasks that can capture coherence-related features. 

We experiment with the following set of sentence-level tasks that are relevant to discourse coherence: 1) the subject number (SubjNum) task that detects the number of the subject of the main clause; 2) the object number (ObjNum) task that detects the number of the direct object of the main clause; 3) the coordination inversion (CoordInv) task that contains sentences consisting of two coordinate clauses, where the two clauses are inverted in half of the sentences and kept intact in the other half (the task is to detect whether a sentence is modified or not); 4) the corrupt agreement (CorruptAgr) task where sentences are corrupted by inverting the verb number (the task is to identify corrupted sentences).

Tasks 1, 2 and 4 align with Centering theory as they probe for subject and object relevant information; the theory suggests that subject and object roles are indicators of entity salience. On the other hand, task 3 tests whether the models can capture intra-sentential coherence. For these tasks, we use the datasets from \citet{conneau-etal-2018-cram} (tasks $1$,$2$ and $3$) and \citet{linzen-etal-2016-assessing} (task $4$).
\vspace{-0.2cm}
\paragraph{Probing model} We adopt the SentEval framework of~\citet{conneau-etal-2018-cram}. 
Our probing model consists of an MLP layer over model sentence representations, followed by sigmoid non-linearity. We use the same training parameters as~\citet{conneau-etal-2018-cram}.\footnote{\url{https://github.com/facebookresearch/SentEval}}

\vspace{-0.2cm}
\paragraph{Results} Table~\ref{table4} presents the results.\footnote{Egrid$_{cnn}$ is based on entity transitions across sentences and therefore we cannot probe sentence representations.}  
Overall, we observe that models are better at detecting SubjNum, and ObjNum (accuracy of at least $61$\% for all models except LC which is the odd one out) compared to CorruptAgr and CoordInv, with the last two being particularly challenging for most models (minimum accuracy of $53$\% excluding LC). For SubjNum and ObjNum the models can find hints in words other than the target word (as the majority of nouns in a sentence tend to have the same number, with $75.9$\% of SubjNum test sentences and $78.7$\% of ObjNum ones containing nouns of the same number in the same sentence~\cite{conneau-etal-2018-cram}).
On the other hand, CorruptAgr examples are longer and with more syntactic variations and require the models to detect the dependency between verbs and their subjects. CoordInv is also a difficult task for the models particularly since they are pre-trained on the WSJ to focus on the order of sentences, not clauses.

Across all tasks, we find that LCD$_{bert}$ achieves the best performance, 
outperforming all other approaches. 
We note, however, that LCD$_{bert}$ does not fine-tune its sentence representations during coherence training in the WSJ but they are rather fixed and based on the average of BERT-based word embeddings (Section \ref{models}). This means the probing model fine-tunes averaged BERT-based word embeddings rather than actual sentence parameters from the LCD coherence model. 
Therefore, the level of performance observed is not representative of the maximum performance coherence models can achieve on these tasks.\footnote{Nevertheless, we observe that LCD$_{bert}$ outperforms the best reported result on CoordInv by \citet{conneau-etal-2018-cram}.} 
We surmise that the comparatively lower performance  observed with MTL$_{bert}$ and STL$_{bert}$ (whose sentence representations are fine-tuned during coherence training) is due to their coherence training objective. The models are optimized on the binary discrimination task, i.e. learning to rank a well-organized document higher than its permuted counterparts. This is an overly simplistic approach to coherence modeling that may be making models (and their representations) more susceptible to losing useful linguistic information.  Having said that, though, MTL$_{bert}$, that has direct training signal with respect to the words' grammatical roles, is able to alleviate this issue to an extent and is the next best performing model on SubjNum, ObjNum and CorruptAgr.

Across tasks, LC is the odd one out, and the worst performing model. This can be explained partly by its comparatively lower performance on the simpler binary discrimination task (Table~\ref{table1}) and partly by the simplicity of the approach: LC utilizes no attention mechanism as the MTL and STL family of models do, nor has expressive enough transformations as LCD$_{rnnlm}$ does.

\section{Discussion} 

Our evaluation experiments on two coherence datasets reveal that RNN- or EGrid-based coherence models are able to detect syntactic alterations that undermine coherence, but are less effecient at detecting semantic ones even after fine-tuning on the latter. 
We furthermore find that they particularly struggle with recognizing minor lexical changes even if they result in implausible meaning and resolving pronominal references. 
On the other hand, these models are particularly good at detecting cases where a prefix is inserted or the subject pronoun is substituted with a lexical item, suggesting that they are capable of capturing the relevant syntactic patterns and do not solely rely on positional features.  
We find that the best performing model overall is LCD$_{bert}$ which does not use an RNN sentence encoder but rather builds sentence representations by averaging BERT embeddings then utilizes a number of linear transformations over adjacent sentences to facilitate learning richer representations.  

Our probing experiments reveal that models are better at encoding information regarding subject and object number 
followed by verb number (CorruptAgr). These probing tasks align with Centering theory as they probe for subject and object relevant information. The task that tests for knowledge on coordination inversion is the lowest performing one overall, suggesting that there is little capacity at capturing information related to intra-sentential coherence. Excluding LCD$_{bert}$, MTL$_{bert}$ is the best performing model; nevertheless, there is still scope for substantial improvement across all probing tasks and particularly on CoordInv and CorruptAgr.

\section{Conclusion}

We systematically studied how well current models of coherence can capture aspects of text implicated in discourse organisation. We devised datasets of various kinds of incoherence and examined model susceptibility to syntactic and semantic alterations. Our results demonstrate the models are robust with respect to corrupted syntactic patterns, prefix insertions and lexical substitutions. However, they fall short in capturing rhetorical and semantic corruptions, lexical perturbations and corrupt pronouns. We furthermore find that discourse embedding space encodes subject and object relevant information; however, there is scope for substantial improvement in terms of encoding linguistic properties relevant to discourse coherence. Experiments on coordination inversion further suggest that current models have little capacity at encoding information related to intra-sentential coherence.

We hope this study shall provide further insight into how to frame the task of coherence modeling and improve model performance further. Finally, we make our datasets publicly available for researchers to use to test coherence models.

\bibliography{emnlp2020}

\begin{thebibliography}{35}
\expandafter\ifx\csname natexlab\endcsname\relax\def\natexlab#1{#1}\fi

\bibitem[{Barzilay and Lapata(2005)}]{Barzilay:2005}
Regina Barzilay and Mirella Lapata. 2005.
\newblock \href {https://www.aclweb.org/anthology/P05-1018} {Modeling local
  coherence: An entity-based approach}.
\newblock In \emph{Proceedings of the 43rd Annual Meeting on Association for
  Computational Linguistics}, pages 141--148. Association for Computational
  Linguistics.

\bibitem[{Barzilay and Lapata(2008)}]{Barzilay2008}
Regina Barzilay and Mirella Lapata. 2008.
\newblock \href {https://www.aclweb.org/anthology/J08-1001.pdf} {Modeling local
  coherence: An entity-based approach}.
\newblock \emph{Computational Linguistics}, 3(1):1--34.

\bibitem[{Barzilay and Lee(2004)}]{barzilay-lee-2004-catching}
Regina Barzilay and Lillian Lee. 2004.
\newblock \href {https://www.aclweb.org/anthology/N04-1015} {Catching the
  drift: Probabilistic content models, with applications to generation and
  summarization}.
\newblock In \emph{Proceedings of the Human Language Technology Conference of
  the North {A}merican Chapter of the Association for Computational
  Linguistics: {HLT}-{NAACL} 2004}, pages 113--120, Boston, Massachusetts, USA.
  Association for Computational Linguistics.

\bibitem[{Burstein et~al.(2010)Burstein, Tetreault, and
  Andreyev}]{Burstein2010}
Jill Burstein, Joel Tetreault, and Slava Andreyev. 2010.
\newblock \href {https://www.aclweb.org/anthology/N10-1099} {Using entity-based
  features to model coherence in student essays}.
\newblock In \emph{Human Language Technologies: The 2010 Annual Conference of
  the North American Chapter of the Association for Computational Linguistics},
  pages 681--684. Association for Computational Linguistics.

\bibitem[{Clark and Manning(2016)}]{clark2016deep}
Kevin Clark and Christopher~D. Manning. 2016.
\newblock \href {https://www.aclweb.org/anthology/D16-1245} {Deep reinforcement
  learning for mention-ranking coreference models}.
\newblock In \emph{Proceedings of the 2016 Conference on Empirical Methods in
  Natural Language Processing}, pages 2256--2262, Austin, Texas. Association
  for Computational Linguistics.

\bibitem[{Conneau et~al.(2018)Conneau, Kruszewski, Lample, Barrault, and
  Baroni}]{conneau-etal-2018-cram}
Alexis Conneau, German Kruszewski, Guillaume Lample, Lo{\"\i}c Barrault, and
  Marco Baroni. 2018.
\newblock \href {https://www.aclweb.org/anthology/P18-1198} {What you can cram
  into a single {\$}{\&}!{\#}* vector: Probing sentence embeddings for
  linguistic properties}.
\newblock In \emph{Proceedings of the 56th Annual Meeting of the Association
  for Computational Linguistics (Volume 1: Long Papers)}, pages 2126--2136,
  Melbourne, Australia. Association for Computational Linguistics.

\bibitem[{Cui et~al.(2018)Cui, Li, Chen, and Zhang}]{Cui2018}
Baiyun Cui, Yingming Li, Ming Chen, and Zhongfei Zhang. 2018.
\newblock \href {https://www.aclweb.org/anthology/D18-1465} {Deep attentive
  sentence ordering network}.
\newblock In \emph{Proceedings of the 2018 Conference on Empirical Methods in
  Natural Language Processing}, pages 4340--4349. Association for Computational
  Linguistics.

\bibitem[{Devlin et~al.(2019)Devlin, Chang, Lee, and
  Toutanova}]{devlin2018bert}
Jacob Devlin, Ming-Wei Chang, Kenton Lee, and Kristina Toutanova. 2019.
\newblock \href {https://www.aclweb.org/anthology/N19-1423} {{BERT}:
  Pre-training of deep bidirectional transformers for language understanding}.
\newblock In \emph{Proceedings of the 2019 Conference of the North {A}merican
  Chapter of the Association for Computational Linguistics: Human Language
  Technologies, Volume 1 (Long and Short Papers)}, pages 4171--4186,
  Minneapolis, Minnesota. Association for Computational Linguistics.

\bibitem[{Elsner et~al.(2007)Elsner, Austerweil, and
  Charniak}]{elsner-etal-2007-unified}
Micha Elsner, Joseph Austerweil, and Eugene Charniak. 2007.
\newblock \href {https://www.aclweb.org/anthology/N07-1055} {A unified local
  and global model for discourse coherence}.
\newblock In \emph{Human Language Technologies 2007: The Conference of the
  North {A}merican Chapter of the Association for Computational Linguistics;
  Proceedings of the Main Conference}, pages 436--443, Rochester, New York.
  Association for Computational Linguistics.

\bibitem[{Elsner and Charniak(2011)}]{Elsner2011}
Micha Elsner and Eugene Charniak. 2011.
\newblock \href {https://www.aclweb.org/anthology/P11-2022} {Extending the
  entity grid with entity-specific features}.
\newblock In \emph{Proceedings of the 49th Annual Meeting of the Association
  for Computational Linguistics: Human Language Technologies}, pages 125--129.
  Association for Computational Linguistics.

\bibitem[{Farag and Yannakoudakis(2019)}]{farag-yannakoudakis-2019-multi}
Youmna Farag and Helen Yannakoudakis. 2019.
\newblock \href {https://www.aclweb.org/anthology/P19-1060} {Multi-task
  learning for coherence modeling}.
\newblock In \emph{Proceedings of the 57th Annual Meeting of the Association
  for Computational Linguistics}, pages 629--639, Florence, Italy. Association
  for Computational Linguistics.

\bibitem[{Feng et~al.(2014)Feng, Lin, and Hirst}]{Feng2014}
Vanessa~Wei Feng, Ziheng Lin, and Graeme Hirst. 2014.
\newblock \href {https://www.aclweb.org/anthology/C14-1089} {The impact of deep
  hierarchical discourse structures in the evaluation of text coherence}.
\newblock In \emph{Proceedings of COLING 2014, the 25th International
  Conference on Computational Linguistics: Technical Papers}, pages 940--949.
  Dublin City University and Association for Computational Linguistics.

\bibitem[{Filippova and Strube(2007)}]{filippova-strube-2007-extending}
Katja Filippova and Michael Strube. 2007.
\newblock \href {https://www.aclweb.org/anthology/W07-2321} {Extending the
  entity-grid coherence model to semantically related entities}.
\newblock In \emph{Proceedings of the Eleventh {E}uropean Workshop on Natural
  Language Generation ({ENLG} 07)}, pages 139--142, Saarbr{\"u}cken, Germany.
  DFKI GmbH.

\bibitem[{Grosz et~al.(1995)Grosz, Weinstein, and Joshi}]{Grosz1995}
Barbara~J. Grosz, Scott Weinstein, and Aravind~K. Joshi. 1995.
\newblock \href {https://www.aclweb.org/anthology/J95-2003} {Centering: A
  framework for modeling the local coherence of discourse}.
\newblock \emph{Computational Linguistics}, 21(2).

\bibitem[{Guinaudeau and Strube(2013)}]{Guinaudeau2013}
Camille Guinaudeau and Michael Strube. 2013.
\newblock \href {https://www.aclweb.org/anthology/P13-1010} {Graph-based local
  coherence modeling}.
\newblock In \emph{Proceedings of the 51st Annual Meeting of the Association
  for Computational Linguistics (Volume 1: Long Papers)}, pages 93--103.
  Association for Computational Linguistics.

\bibitem[{Halliday and Hasan.(1976)}]{halliday}
M.~A.~K. Halliday and Ruqaiya Hasan. 1976.
\newblock \emph{Cohesion in English. Longman, London.}

\bibitem[{Hewitt and Manning(2019)}]{hewitt2019structural}
John Hewitt and Christopher~D Manning. 2019.
\newblock A structural probe for finding syntax in word representations.
\newblock In \emph{Proceedings of the 2019 Conference of the North American
  Chapter of the Association for Computational Linguistics: Human Language
  Technologies, Volume 1 (Long and Short Papers)}, pages 4129--4138.

\bibitem[{Joty et~al.(2018)Joty, Mohiuddin, and Tien~Nguyen}]{Joty2018}
Shafiq Joty, Muhammad~Tasnim Mohiuddin, and Dat Tien~Nguyen. 2018.
\newblock \href {https://www.aclweb.org/anthology/P18-1052} {Coherence modeling
  of asynchronous conversations: A neural entity grid approach}.
\newblock In \emph{Proceedings of the 56th Annual Meeting of the Association
  for Computational Linguistics (Volume 1: Long Papers)}, pages 558--568.
  Association for Computational Linguistics.

\bibitem[{Kingma and Ba(2015)}]{KingmaB14}
Diederik~P. Kingma and Jimmy Ba. 2015.
\newblock \href {http://arxiv.org/abs/1412.6980} {Adam: {A} method for
  stochastic optimization}.
\newblock In \emph{3rd International Conference on Learning Representations,
  {ICLR} 2015, San Diego, CA, USA, May 7-9, 2015, Conference Track
  Proceedings}.

\bibitem[{Lai and Tetreault(2018)}]{Lai2018}
Alice Lai and Joel Tetreault. 2018.
\newblock \href {https://www.aclweb.org/anthology/W18-5023} {Discourse
  coherence in the wild: A dataset, evaluation and methods}.
\newblock In \emph{Proceedings of the 19th Annual SIGdial Meeting on Discourse
  and Dialogue}, pages 214--223. Association for Computational Linguistics.

\bibitem[{Lapata and Barzilay(2005)}]{Lapata:2005:AET:1642293.1642467}
Mirella Lapata and Regina Barzilay. 2005.
\newblock \href {http://dl.acm.org/citation.cfm?id=1642293.1642467} {Automatic
  evaluation of text coherence: Models and representations}.
\newblock In \emph{Proceedings of the 19th International Joint Conference on
  Artificial Intelligence}, IJCAI'05, pages 1085--1090, San Francisco, CA, USA.
  Morgan Kaufmann Publishers Inc.

\bibitem[{Li and Jurafsky(2017)}]{Li2017}
Jiwei Li and Dan Jurafsky. 2017.
\newblock \href {https://www.aclweb.org/anthology/D17-1019} {Neural net models
  of open-domain discourse coherence}.
\newblock In \emph{Proceedings of the 2017 Conference on Empirical Methods in
  Natural Language Processing}, pages 198--209. Association for Computational
  Linguistics.

\bibitem[{Lin et~al.(2011)Lin, Ng, and Kan}]{lin-etal-2011-automatically}
Ziheng Lin, Hwee~Tou Ng, and Min-Yen Kan. 2011.
\newblock \href {https://www.aclweb.org/anthology/P11-1100} {Automatically
  evaluating text coherence using discourse relations}.
\newblock In \emph{Proceedings of the 49th Annual Meeting of the Association
  for Computational Linguistics: Human Language Technologies}, pages 997--1006,
  Portland, Oregon, USA. Association for Computational Linguistics.

\bibitem[{Linzen et~al.(2016)Linzen, Dupoux, and
  Goldberg}]{linzen-etal-2016-assessing}
Tal Linzen, Emmanuel Dupoux, and Yoav Goldberg. 2016.
\newblock \href {https://doi.org/10.1162/tacl_a_00115} {Assessing the ability
  of {LSTM}s to learn syntax-sensitive dependencies}.
\newblock \emph{Transactions of the Association for Computational Linguistics},
  4:521--535.

\bibitem[{Liu et~al.(2019)Liu, Gardner, Belinkov, Peters, and
  Smith}]{liu2019linguistic}
Nelson~F Liu, Matt Gardner, Yonatan Belinkov, Matthew Peters, and Noah~A Smith.
  2019.
\newblock Linguistic knowledge and transferability of contextual
  representations.
\newblock \emph{arXiv preprint arXiv:1903.08855}.

\bibitem[{Logeswaran et~al.(2018)Logeswaran, Lee, and Radev}]{Logeswaran2018}
Lajanugen Logeswaran, Honglak Lee, and Dragomir~R. Radev. 2018.
\newblock Sentence ordering and coherence modeling using recurrent neural
  networks.
\newblock In \emph{AAAI}, pages 5285--5292. AAAI Press.

\bibitem[{Louis and Nenkova(2012)}]{louis-nenkova-2012-coherence}
Annie Louis and Ani Nenkova. 2012.
\newblock \href {https://www.aclweb.org/anthology/D12-1106} {A coherence model
  based on syntactic patterns}.
\newblock In \emph{Proceedings of the 2012 Joint Conference on Empirical
  Methods in Natural Language Processing and Computational Natural Language
  Learning}, pages 1157--1168, Jeju Island, Korea. Association for
  Computational Linguistics.

\bibitem[{Mann and Thompson(1988)}]{mann1988rhetorical}
William~C Mann and Sandra~A Thompson. 1988.
\newblock Rhetorical structure theory: Toward a functional theory of text
  organization.
\newblock \emph{Text Interdisciplinary Journal for the Study of Discourse},
  pages 243--281.

\bibitem[{Moon et~al.(2019)Moon, Mohiuddin, Joty, and
  Xu}]{moon-etal-2019-unified}
Han~Cheol Moon, Tasnim Mohiuddin, Shafiq Joty, and Chi Xu. 2019.
\newblock \href {https://www.aclweb.org/anthology/D19-1231} {{A Unified Neural
  Coherence Model}}.
\newblock In \emph{Proceedings of the 2019 Conference on Empirical Methods in
  Natural Language Processing and the 9th International Joint Conference on
  Natural Language Processing (EMNLP-IJCNLP)}, pages 2262--2272, Hong Kong,
  China. Association for Computational Linguistics.

\bibitem[{Mostafazadeh et~al.(2016)Mostafazadeh, Chambers, He, Parikh, Batra,
  Vanderwende, Kohli, and Allen}]{mostafazadeh-etal-2016-corpus}
Nasrin Mostafazadeh, Nathanael Chambers, Xiaodong He, Devi Parikh, Dhruv Batra,
  Lucy Vanderwende, Pushmeet Kohli, and James Allen. 2016.
\newblock \href {https://doi.org/10.18653/v1/N16-1098} {A corpus and cloze
  evaluation for deeper understanding of commonsense stories}.
\newblock In \emph{Proceedings of the 2016 Conference of the North {A}merican
  Chapter of the Association for Computational Linguistics: Human Language
  Technologies}, pages 839--849, San Diego, California. Association for
  Computational Linguistics.

\bibitem[{Pennington et~al.(2014)Pennington, Socher, and
  Manning}]{pennington2014glove}
Jeffrey Pennington, Richard Socher, and Christopher~D. Manning. 2014.
\newblock \href {http://www.aclweb.org/anthology/D14-1162} {{GloVe}: Global
  vectors for word representation}.
\newblock In \emph{Empirical Methods in Natural Language Processing (EMNLP)},
  pages 1532--1543.

\bibitem[{Somasundaran et~al.(2014)Somasundaran, Burstein, and
  Chodorow}]{somasundaran-etal-2014-lexical}
Swapna Somasundaran, Jill Burstein, and Martin Chodorow. 2014.
\newblock \href {https://www.aclweb.org/anthology/C14-1090} {Lexical chaining
  for measuring discourse coherence quality in test-taker essays}.
\newblock In \emph{Proceedings of {COLING} 2014, the 25th International
  Conference on Computational Linguistics: Technical Papers}, pages 950--961,
  Dublin, Ireland. Dublin City University and Association for Computational
  Linguistics.

\bibitem[{Soricut and Marcu(2006)}]{soricut-marcu-2006-discourse}
Radu Soricut and Daniel Marcu. 2006.
\newblock \href {https://www.aclweb.org/anthology/P06-2103} {Discourse
  generation using utility-trained coherence models}.
\newblock In \emph{Proceedings of the {COLING}/{ACL} 2006 Main Conference
  Poster Sessions}, pages 803--810, Sydney, Australia. Association for
  Computational Linguistics.

\bibitem[{Tien~Nguyen and Joty(2017)}]{Dat2017}
Dat Tien~Nguyen and Shafiq Joty. 2017.
\newblock \href {https://www.aclweb.org/anthology/P17-1121} {A neural local
  coherence model}.
\newblock In \emph{Proceedings of the 55th Annual Meeting of the Association
  for Computational Linguistics (Volume 1: Long Papers)}, pages 1320--1330.
  Association for Computational Linguistics.

\bibitem[{Xu et~al.(2019)Xu, Saghir, Kang, Long, Bose, Cao, and
  Cheung}]{xu-etal-2019-cross}
Peng Xu, Hamidreza Saghir, Jin~Sung Kang, Teng Long, Avishek~Joey Bose,
  Yanshuai Cao, and Jackie Chi~Kit Cheung. 2019.
\newblock \href {https://doi.org/10.18653/v1/P19-1067} {A cross-domain
  transferable neural coherence model}.
\newblock In \emph{Proceedings of the 57th Annual Meeting of the Association
  for Computational Linguistics}, pages 678--687, Florence, Italy. Association
  for Computational Linguistics.

\end{thebibliography}
\bibliographystyle{acl_natbib}

\end{document}